\documentclass[sigconf]{acmart}
\AtBeginDocument{%
  \providecommand\BibTeX{{%
    \normalfont B\kern-0.5em{\scshape i\kern-0.25em b}\kern-0.8em\TeX}}}

\copyrightyear{2022}
\acmYear{2022}
\setcopyright{acmcopyright}\acmConference[ACM MobiCom '21]{The 27th Annual International Conference on Mobile Computing and Networking}{January 31-February 4, 2022}{New Orleans, LA, USA}
\acmBooktitle{The 27th Annual International Conference on Mobile Computing and Networking (ACM MobiCom '21), January 31-February 4, 2022, New Orleans, LA, USA}
\acmPrice{15.00}
\acmDOI{10.1145/3447993.3483249}
\acmISBN{978-1-4503-8342-4/22/01}

\usepackage{multirow}

\usepackage{amsmath}
\usepackage{enumitem}
\setlist{leftmargin=3mm}
\usepackage{amssymb}

\usepackage{algorithm}
\usepackage{algorithmic}
\usepackage{url}
\usepackage{threeparttable}

\usepackage{hyperref}
\hypersetup{breaklinks=true}
\usepackage[hyphenbreaks]{breakurl}

\usepackage{graphicx}
\usepackage{color} 

\begin{document}

\title{LegoDNN: Block-grained Scaling of Deep Neural Networks for Mobile Vision}

\author{Rui Han}
\email{hanrui@bit.edu.cn}
\affiliation{%
  \institution{Beijing Institute of Technology}
  \city{Beijing}
  \country{P.R. China}}

\author{Qinglong Zhang}
\email{3120211050@bit.edu.cn}
\affiliation{%
  \institution{Beijing Institute of Technology}
  \city{Beijing}
  \country{P.R. China}}

\author{Chi Harold~Liu}
\email{chiliu@bit.edu.cn}
\affiliation{%
  \institution{Beijing Institute of Technology}
  \city{Beijing}
  \country{P.R. China}}

\author{Guoren Wang}
\email{wanggr@bit.edu.cn}
\affiliation{%
  \institution{Beijing Institute of Technology}
  \city{Beijing}
  \country{P.R. China}}

\author{Jian Tang}
\email{tangjian22@midea.com}
\affiliation{%
  \institution{Midea Group}
   \city{Shanghai}
   \country{P.R. China}}

\author{Lydia Y.~Chen}
\email{lydiaychen@ieee.org}
\affiliation{%
  \institution{TU Delft}
   \city{Delft}
   \country{Netherlands}}

\renewcommand{\shortauthors}{Han and Zhang, et al.}

\begin{abstract}
Deep neural networks (DNNs) have become ubiquitous techniques in mobile and embedded systems for applications such as image/object recognition and classification. The trend of executing multiple DNNs simultaneously exacerbate the existing limitations of meeting stringent latency/accuracy requirements on resource constrained mobile devices. The prior art sheds light on exploring the accuracy-resource tradeoff by scaling the model sizes in accordance to resource dynamics. However, such model scaling approaches face to imminent challenges: (i) large space exploration of model sizes, and (ii) prohibitively long training time for different model combinations. In this paper, we present LegoDNN, a lightweight, block-grained scaling solution for running multi-DNN workloads in mobile vision systems. LegoDNN guarantees short model training times by only extracting and training a small number of common blocks (e.g. 5 in VGG and 8 in ResNet) in a DNN. At run-time, LegoDNN optimally combines the descendant models of these blocks to maximize accuracy under specific resources and latency constraints, while reducing switching overhead via smart block-level scaling of the DNN. We implement LegoDNN in TensorFlow Lite and extensively evaluate it against state-of-the-art techniques (FLOP scaling, knowledge distillation and model compression) using a set of 12 popular DNN models. Evaluation results show that LegoDNN provides 1,296x to 279,936x more options in model sizes without increasing training time, thus achieving as much as 31.74\% improvement in inference accuracy and 71.07\% reduction in scaling energy consumptions.
\end{abstract}



\maketitle

\section{Introduction} \label{Section: Introduction}

Deep neural networks (DNNs) have become ubiquitous in computer vision applications~\cite{howard2017mobilenets}, spanning from image recognition~\cite{simonyan2014very,han2019slimml} to object detection~\cite{EdgeInference19,riazi2018chameleon} to video analytics~\cite{jiang2018mainstream}.
Today's mobile systems run multiple vision related applications which are based on DNN models~\cite{likamwa2015starfish,NestDNN18,bateni2020neuos}. For instance, mobile applications of object detection and image classification are actively executed on all the images taken. As a result, inference jobs of two-DNN concurrently run on mobile devices. Such an example can be extended further to more complex scenarios, where more DNN based applications are involved.
Typically, DNNs consume large amounts of resources and require minimum processing latency and maximum inference accuracy. The stringent computation, memory, and power constraints of mobile devices (e.g. smart phones, tablets, and embedded devices) hinder the performance of such DNNs.
The challenging of coping with resource limitation is further exacerbated when encountering events such as applications start and exist at runtime.

\begin{figure}[htp]
\centering
  \includegraphics[scale=0.47]{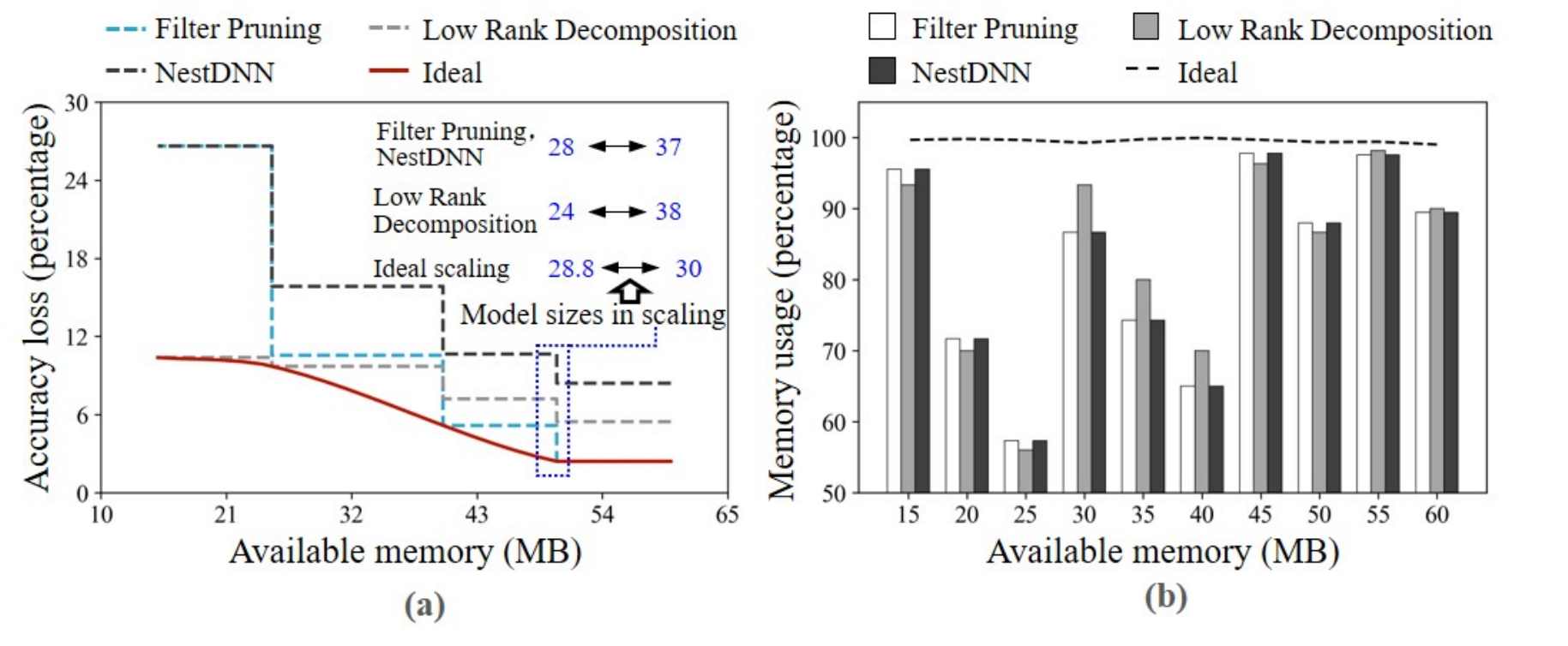}\\
  \caption{Accuracy losses and resource utilization of model-grained scaling}
  \label{Fig: ScalingExample}
\end{figure}

\textbf{Limitations of model-grained scaling}. The key to achieve the full promise of accurate DNNs is to effectively scale their sizes in correspondence to the dynamics of available resource.
Existing layer removing techniques~\cite{han2015deep,oh2018portable,li2018jalad,yang2017designing,reagen2016minerva}  such as filter pruning require re-training of a model to mitigate its severe accuracy loss once its size is changed.
To provide online scaling, these techniques need to first transform a DNN model into several descendant models of smaller sizes offline, and then select one of them to upgrade or downgrade the model size at run-time.
However, \emph{these models only provide a coarse-grained differentiation of resources, often incurring large accuracy losses (Figure \ref{Fig: ScalingExample}(a)) and leaving considerably proportions of resource unused (Figure \ref{Fig: ScalingExample}(b))}.
An ideal scaling that fully uses the available resources requires a dauntingly large number of descendant models, which gives rise to two major challenges in practice.

\emph{Fast generation of descendant models}. In structured pruning techniques such as filter pruning~\cite{li2016pruning} and low rank decomposition~\cite{kim2015compression}, re-training of a descendant model takes most of its generation time. When handling large models and datasets, the re-training of one model takes hours to complete even using powerful GPU servers. The ideal scaling requires at least hundreds of models and the whole training process takes several months.
The \textbf{first challenge}, therefore, is how to construct a large number (e.g. 1 million) of descendant models using a moderate amount of time such as a few hours.

\emph{Low overhead and high accuracy in model scaling}. At run-time, massive descendant models also cause frequent model switches in scaling, and the page-in and page-out overheads of an entire model cause considerable energy consumptions.
With large exploration space of model sizes, finding which model capable of maximizing inference accuracy under resource dynamics and latency constraint is not trivial. This is because two descendant models of the \emph{same} size may have \emph{different} inference accuracies. The \textbf{second challenge} now requires us to depart from the model-oriented view of scaling and investigate how to manage model scaling at a finer granularity.

In this paper, we present LegoDNN, a framework that is designed fundamentally based on the block structure (e.g. ResBlock in ResNet~\cite{resnet}) widely exists in DNNs of today's mobile vision systems.
The key idea of LegoDNN is to introduce a \textbf{training environment} to represent the input and output channels of a block in a DNN. This environment enables the re-training of the block's descendant/compressed blocks independent of other parts of the DNN. At run-time, the environment supports the dynamic scaling of a DNN by just switching a part of descendant blocks in it and avoids re-training after model size changes.
In doing so, LegoDNN addresses the limitations of existing model-grained techniques in both time-consuming model re-training and coarse-grained model scaling. In particular, the contributions of this paper are as follows:

\begin{itemize}[noitemsep,nolistsep]
\item \textbf{Fast generation of large scaling space via block re-training}. LegoDNN proposes a novel block-level approach that extracts blocks from an entire DNN and re-trains their descendant blocks independent of the whole network. This allows the training to be completed fast by just generating a few blocks for a DNN, while the combination of these blocks brings massive options of descendant models.

\item \textbf{Fine-grained DNN scaling via block switching}. LegoDNN employs a runtime scalar to select the optimal combination of blocks for maximum accuracy. Comparing to layer removing techniques such as filter pruning, our approach only switches the blocks that do not exist in the new combination, thus providing online latency-accuracy trade-offs without re-training of the entire network and also reducing scaling overheads.

\item \textbf{Implementation and Evaluation}. We design and develop TensorFlow applications, which visualize the inference process of LegoDNN and enhance the usability. We conduct exhausted evaluation against the state-of-the-art techniques, i.e. FLOP scaling, knowledge distillation and model compression, using 12 representative DNNs in both single and multi-DNN execution scenarios.
\end{itemize}

\textbf{Summary of experimental results.}
\emph{(i) Extensible in terms of architecture}. We fully implement LegoDNN on both smart phones (Huawei Mate20 X and Xiaomi 5S) and embedded devices (Raspberry Pi 4B and Jetson TX2).
\emph{(ii) Efficient block training and scaling}. We extensively compare LegoDNN to state-of-the-art techniques, and find that LegoDNN provides 1,296x to 279,936x more optional descendant models while using 24\% of model generation time.
     Using these blocks, our approach achieves 31.74\% increase in accuracy under the same resource budgets.
    LegoDNN also reduces runtime model switching energies by 71.07\%.
\emph{(iii) Multi-DNN scenario}. We ensure that our system can trade-off and balance resources among multiple DNNs by testing LegoDNN under different loads and scheduling policies.
\emph{(iv) Applicability to DNNs and model compression techniques.}
We demonstrate the applicability of our approach in seven categories of prevalent DNNs proposed in recent years~\cite{khan2020survey}. We also integrate our approach with two data-independent model compression, three data-dependent techniques, and one quantized based solution.

\section{Motivation} \label{Section: Motivation}

In this section, we lay out several observations to understand the block structure in DNN scaling, and gain insights on the limitation of model-grained scaling.

\begin{figure*}[!t]
\centering
  \includegraphics[scale=0.33]{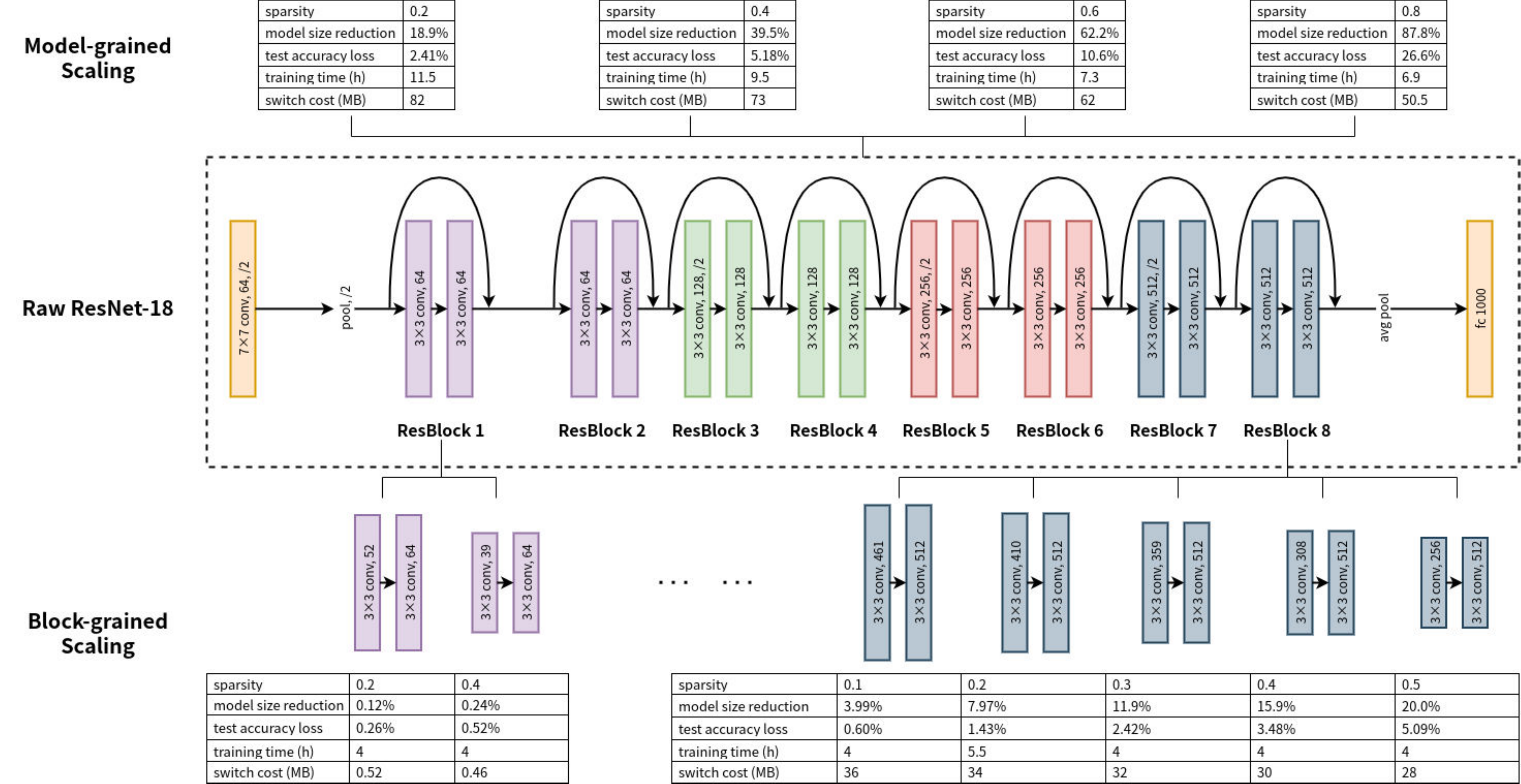}\\
  \caption{A motivation example to understand the block structure in DNN scaling}
  \label{Fig: Motivation}
\end{figure*}

\textbf{Limitation 1: limited exploration of model choices}.
Traditional model-grained scaling techniques treat a DNN as a whole and transform it into several descendant models of different sparsities (model sizes), as shown in Figure \ref{Fig: Motivation}'s upper side. However, the large discrepancy among these models often results in insufficient usage of available resources at run-time.

\textbf{Blocks in DNN}. Thus, in order to achieve a better accuracy-latency tradeoff in a finer granularity, a more detail DNN structure information, such as network block, must be transparent to the mobile vision system.
Conceptually, a \textbf{block} \emph{is one part of a network that represents its repeating pattern of layers}.
In Figure \ref{Fig: Motivation}'s example, ResNet18 consists of 8 ResBlocks and the reductions of model sizes are much smaller in these descendant blocks.
This means block-grained scaling can make the selection space of model size \emph{two to five orders of magnitude larger}.
For instance, in three popular DNNs (VGG16~\cite{VGG}, YOLOv3~\cite{YOLO}, and ResNet18~\cite{resnet}), the block-grained scaling provides as much as 7,776 to 1,679,616 model sizes by just transforming the original blocks into 5 to 40 descendant blocks.

\textbf{Research question 1}: \emph{How to enlarge the model exploration spaces through blocks?}

\textbf{Limitation 2: Resource intensive and time consuming training}.
In existing model compression techniques~\cite{li2016pruning,NestDNN18,kim2015compression}, the re-training of a descendant model boosts its accuracy but takes most of its generation time. When handling large datasets such as ImageNet or COCO2017, the generation of a block takes several hours. Hence in terms of absolute number of descendant blocks, all these techniques provide limited choices.

\textbf{Research question 2}: \emph{How to accelerate the training of block-grain models?}

\textbf{Limitation 3: Focus on off-line structure pruning}.
Given large search space of descendant models, an efficient scaling approach is required to quickly find the optimal combination of blocks under run-time resource dynamics.
However, today's DNNs have various architectures and the blocks in a DNN consist of different layers, which have different \emph{computational costs} and \emph{memory footprints}. More interestingly, blocks at different layers of the network have \emph{considerably different influences on accuracy}.
Hence when downgrading/upgrading the model size, the scaling approach should first scale down/up a block that causes the smallest/largest accuracy losses per unit of model size reduction/increase.

\textbf{Research question 3}: \emph{how to optimally assemble descendant block on the fly?}

\section{Design of LegoDNN} \label{Section: Method}

To enable the block-grained DNN training and scaling, two basic research questions need to be answered in LegoDNN: 1) how to efficiently extract, train, and profile the descendant blocks of different DNNs (Section \ref{Section: Offline Block Generation}), and 2) in online scaling, how to find the optimal combination of these blocks and efficiently switch them (Section \ref{Section: Online Optimization}).

\subsection{Overview} \label{Section: Overview}

\begin{figure*}[htp]
\centering
  \includegraphics[scale=0.39]{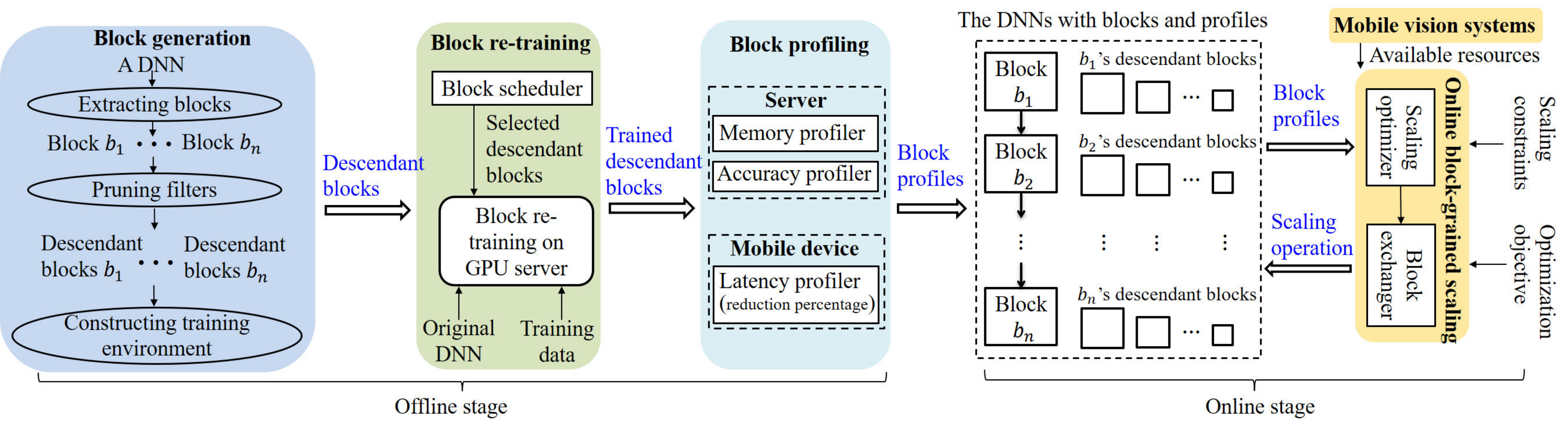}\\
  \caption{LegoDNN architecture}
  \label{Fig: Overview}
\end{figure*}

Figure \ref{Fig: Overview} illustrates the architecture of LegoDNN, which is split into an offline stage and an online stage.

The offline stage consists of three phases: block generation, block re-training, and block profiling (Section \ref{Section: Offline Block Generation}).
In the \emph{block generation} phase, LegoDNN first exacts blocks from a DNN and applies pruning filter techniques to transform each block into a list of descendant blocks. It then constructs each block's \emph{training environment}. In the following re-training, all its descendant blocks have the same training environment. That is, a descendant block's re-training is independent of other parts of the DNN and can be completed much faster than re-training a descendant model in existing approaches.
This environment is important for the fast scaling of the DNN at run-time, because it allows LegoDNN to incorporate any of these descendant blocks into the network without extra operations.

In the \emph{block re-training} phase, LegoDNN employs a \emph{block scheduler} to manage the descendant blocks in the DNN with consideration of both the available memory and processing power in the GPU server. The scheduler aims to put as many blocks as possible in the memory and train them in parallel, while not delaying the training of these blocks due to the time-sharing of GPU cores.
Each descendant block is re-trained using the standard SGD method, and learns from the original/uncompressed block (Section \ref{Section: Descendant Block Generation and Re-training}).

In the \emph{block profiling} phase (Section \ref{Section: Block Profilers}), given a mobile vision system, a profile is generated for each block or descendant block including its inference accuracy, memory footprint, and inference latency. Note that in latency profiling, LegoDNN measures the \emph{percentage} of reduced latency compared to that of the original block. This is because the latency of processing the same block also depends on the run-time system status, e.g. this latency increases when the system resource is saturated. The profile hence records the reduction percentage in order to be applicable to different system statuses.

The online stage consists of two modules: scaling optimizer and block exchanger (Section \ref{Section: Online Optimization}). The \emph{scaling optimizer} continuously monitors the system status. Once any change of available resources is detected, the optimizer checks up all DNNs' profiles of descendant blocks, and finds the optimal combination of these DNNs that maximizes the inference accuracy under the processing latency constraints.
Using the optimization results, the \emph{block exchanger} identifies each DNN's part of descendant blocks that need to change and only exchange these blocks in DNN scaling.

\begin{figure}[htp]
\centering
  \includegraphics[scale=0.6]{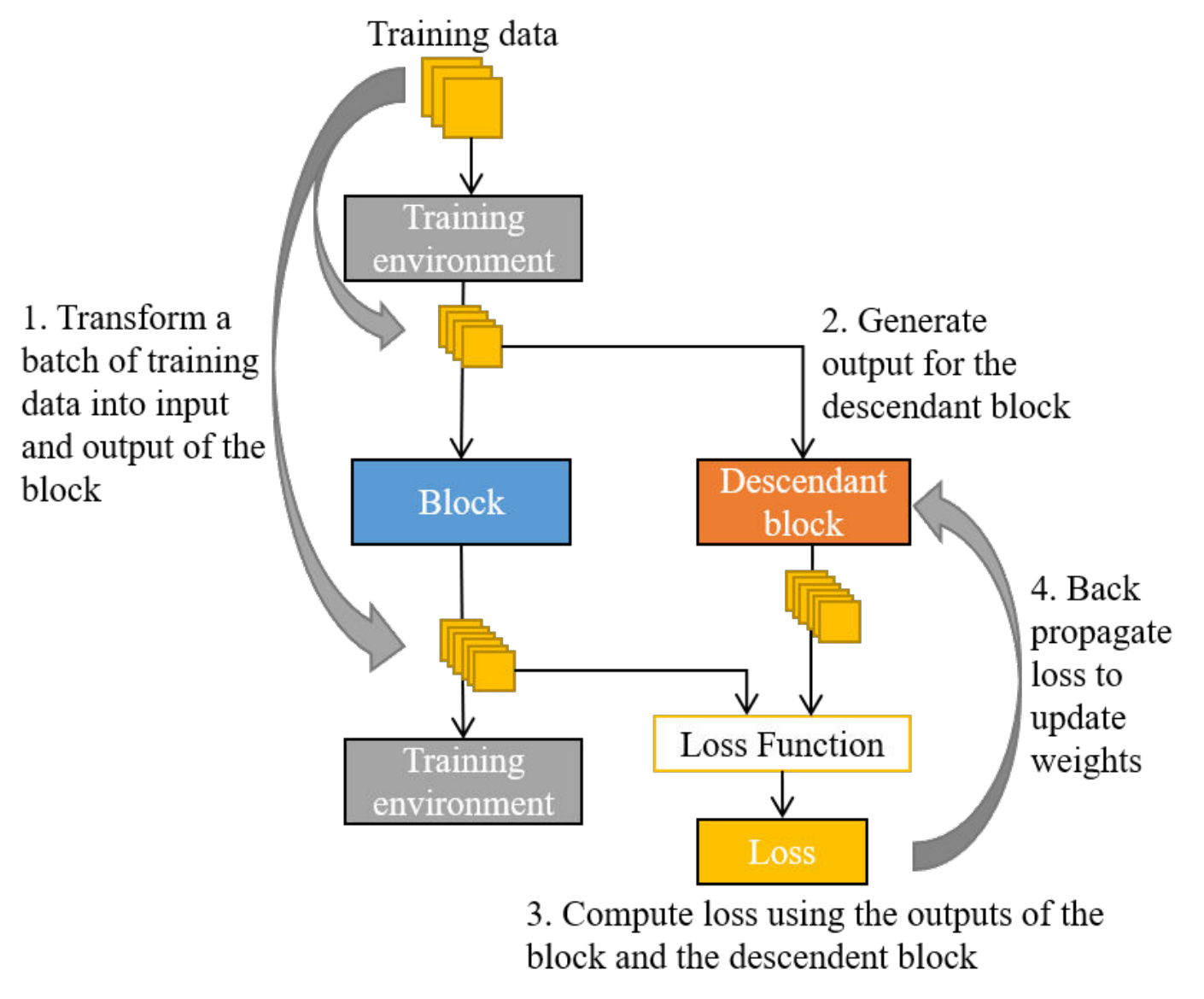}\\
  \caption{Process of re-training of a descendant block}
  \label{Fig: descendantBlockRetrain}
\end{figure}

\subsection{Offline Block Generation and Profiling} \label{Section: Offline Block Generation}

\subsubsection{Descendant Block Generation and Re-training} \label{Section: Descendant Block Generation and Re-training}
\

Given a DNN, the blocks are identified such that these blocks take a majority of computations and memory footprints of the model.
The identification process has two phases. The first phase iteratively gets the shallowest convolutional layer in the DNN, finds all the layers that depend on this layer, and selects these layers to form an \emph{elementary} block.
The second phase combines the \emph{elementary} block to a pre-specified number of blocks. Typically, 5 to 10 blocks are sufficient to provide large scaling space (e.g. 1 million scaling options).

After block identification, the standard pruning filtering technique~\cite{li2016pruning} is then applied to transform each block into a list of descendant blocks, each of them represents one sparsity (one compressed model size) of the original block. Hence in re-training, all descendant blocks have the same training environment and use the original block as the optimization target.
Figure \ref{Fig: descendantBlockRetrain} illustrates the four re-training steps of a descendant block.
At each iteration, step 1 transforms a mini-batch of training data into intermediate data (input and output) of the block.
Step 2 uses the intermediate data to perform inference on the descendant block and generates its output.
Step 3 computes the loss according to the outputs of the original block and the descendant block.
Finally, step 4 back propagates the loss to update the parameters in the descendant block.
The training can be completed quickly for two reasons. First, it only uses the block's intermediate data and avoids the unnecessary computations on other parts of the network. Second, the intermediate data at one iteration can be used to train multiple descendant blocks in parallel.

\subsubsection{Block Profilers}  \label{Section: Block Profilers}
\

The \emph{memory profiler} directly obtains a block's memory footprint according to its model size.
In contrast, obtaining a descendant block's accuracy is not trivial since, in practice, the block may combine with other blocks of different sizes. In DNNs of different model sizes, this block may incur different accuracy losses compared to the original block. For such an issue, the \emph{accuracy profiler} considers $k$ representative model sparsities/sizes in the profiling of each descendant block and calculate the block $b_{i,j}$'s accuracy loss $\mathcal{A}_{i,j}$ as the average accuracy loss of these $k$ profiles.

The \emph{latency profiler} predicts the \textbf{percentage of latency reduction} for each descendant block. This prediction approach is inspired by one key intuition: \emph{a block's processing latency is proportional to its block size, namely its number of parameters}.  Empirical evaluation shows that under different performance interferences, each descendant block has very similar percentages of latency reductions (the difference is less than 5\%).
Let $s_i$ and $s_{i,j}$ be the block sizes of block $b_i$ and its descendant block $b_{i,j}$, the percentage of latency reduction $b_{i,j}$ is calculated as $\mathcal{T}_{i,j}=\frac{s_{i,j}}{s_i}$.
Using $\mathcal{T}_{i,j}$, $b_{i,j}$'s latency can be computed at run-time using two steps. The first step obtains the latency of descendant block $b_{i,v}$ that is currently used by the DNN in the system. This latency represents the system status and the original block $b_i$'s latency can be estimated as $t_i$=$\frac{b_{i,v}}{1-\mathcal{T}_{i,v}}$. The second step then calculates $b_{i,j}$'s latency as $t_{i,j}$=$t_i \cdot (1-\mathcal{T}_{i,j})$.
Note that the profiler can be streamlined (e.g. constructing a regression model) over time as we collect more relationship information between blocks' sizes and their percentages of latency reduction.

\subsection{Online Block-grained Scaling} \label{Section: Online Optimization}

\subsubsection{Scaling Optimizer}
\

\textbf{Single-DNN Optimization}
At run-time, the scaling optimizer finds the optimal combination of (descendant) blocks for a DNN.
The optimization objective is to maximize the DNN's inference accuracy (i.e. minimizing its accuracy loss) under four constraints. Equation \ref{Optimization} defines the optimization objective $o_k$ for the $k$th scaling ($k\geq 1$).

\begin{equation}
\label{Optimization}
\begin{split}
& o_k = \min\limits_{\mathcal{B}} \sum_{i=1}^{n} \sum_{j=0}^{n_i} \mathcal{A}_{i, j} \cdot \mathcal{B}^{(k)}_{i,j} \\
& s.t. \\
& \sum_{i=1}^{n} \sum_{j=0}^{n_i} t_{i,j}  \cdot \mathcal{B}^{(k)}_{i,j} \leq t^{max} \\
& s^{base}-\sum_{i=1}^{n} \sum_{j=0}^{n_i} \mathcal{S}_{i,j}  \cdot \mathcal{B}^{(k)}_{i,j}  \leq s^{max} \\
& \sum_{j=0}^{n_i} \mathcal{B}^{(k)}_{i,j} = 1, \mathcal{B}^{(k)}_{i,j} \in \{0, 1\}\\
\end{split}
\end{equation}

\begin{itemize}
    \item The decision variable $\mathcal{B}^{(k)}_{i,j}$ for a descendant block is either 0 (not selected) or 1 (selected).
    \item The total processing latencies $\sum_{i=1}^{n} \sum_{j=0}^{n_i} t_{i,j} \cdot \mathcal{B}^{(k)}_{i,j}$ of the $n$ selected descendant blocks should be smaller than $t^{max}$ in order to meet the latency constraint.
    \item Let $s^{base}$ be the model size before scaling. The total decreased model size (or the minus of increased model size) in the $n$ blocks is calculated as: $\sum_{i=1}^{n} \sum_{j=0}^{n_i} \mathcal{S}_{i,j} \cdot \mathcal{B}^{(k)}_{i,j}$.
        This constraint requires that after scaling, the model size should be smaller than $s^{max}$ in order to run the model in the available memory.
    \item The block-grained scaling maintains the layer structure of the DNN, and only selects one from each block's descendant blocks. The final constraint states that there only allows one descendant block to be selected for each block: $\forall\ 1 \leq i \leq n, \sum_{j=0}^{n_i} \mathcal{B}^{(k)}_{i,j} = 1$.
\end{itemize}

The optimizer can also address the optimization problem of minimizing latency under the accuracy constraint~\cite{riazi2018chameleon}, by just changing the optimization objective as maximizing the latency reductions: $o_k=\max\limits_{\mathcal{B}} \sum_{i=1}^{n} \sum_{j=0}^{n_i} \mathcal{T}_{i, j} \cdot \mathcal{B}^{(k)}_{i,j}$.

\textbf{{Multi-DNN optimization}}. When considering $m$ DNNs ($m > 1$), the optimization objective can be extended as: \begin{equation}
\label{MultiDNNOptimization}
\begin{split}
o_k = \min \sum_{a=1}^{m} \sum_{i=1}^n \sum_{j=0}^{n_i} \mathcal{A}^{(a)}_{i,j} \cdot \mathcal{B}^{(a,k)}_{i,j}
\end{split}
\end{equation}
where $\mathcal{A}^{(a)}_{i,j}$ represents the accuracy loss of block $b_{i,j}$ in the $a$th DNN, and $\mathcal{B}^{(a,k)}_{i,j}$ denotes whether this block is selected in the $k$th scaling. In Equation \ref{MultiDNNOptimization}, each DNN (i.e. a vision task) has a user-specified latency constraint and its own memory footprint. In optimization, the summarized memory footprints of all $m$ DNNs follow the memory constraint on the mobile device.

The optimization addressed here can be seen as a Integer Linear Programming (ILP) problem because its decision variable $\mathcal{B}^{(k)}_{i,j}$ can only be 0 or 1. We solve this ILP problem using two phases. Phase 1 relaxes the integer constraint of ILP and transforms it into a linear programming (LP) problem $p^*$ that can be quickly solved.
Phase 2 employs the branch and bound method to recursively search the optimal solution of ILP with polynomial time complexity.
The basic idea of this method is to construct a tree that uses $p^*$'s optimal solution $q^*$ as the root node, and recursively traverses its branch nodes until the optimal solution of ILP is found.
We also add an early stopping threshold $\sigma$ such that the search process is completed if the found optimal solution is close enough to the actual optimal solution.
Specifically, for each branch node, the method first rounds down its corresponding solution $q$ to get a feasible solution.
Let $o^{opt}$ be the minimal objective value of all feasible solutions in the traversed branch nodes and $o^{remain}$ be the the optimal objective value of the remaining solutions. The search process completes if $o^{opt} - o^{remain} < \sigma$; otherwise, solution $q$ is splitted into two solutions (namely two new branch nodes), and the method recursively traverses these nodes.

\subsubsection{Block Exchanger}
\

Our block exchanger is designed to reduce high scaling overheads in existing DNN scaling techniques.
First, most of scaling techniques such as \emph{filter pruning} and \emph{low rank decomposition} directly replace the entire DNN to a smaller or larger one, thus may incur large I/O overheads.
NestDNN partially solves this problem by designing a nested structure of descendant model and decreasing the amounts of exchanged pages in model scaling~\cite{NestDNN18}. However, it incurs two extra overheads: (1) the generation of this heavy nested model takes 50\% to 200\% longer training time than the standard filter pruning technique~\cite{li2016pruning}; (2) when some pages are loaded into memory, new layers needs to be constructed using model weights and APIs of deep learning libraries. These operations consume extra resources and time in scaling.

In contrast, our block exchanger provides a lightweight scaling scheme with twofold meanings. First, it only needs to exchange some blocks rather than the whole network, thus saving page-in and page-out overheads like NestDNN.
Second, the newly added blocks can be directly combined into the model without extra operations. This is because in scaling, LegoDNN only exchanges a descendant block to another descendant block belonging to the same original block. In re-training, LegoDNN maintains the same training environment (input and output channels) for both descendant blocks, thereby allowing the switch of them without extra overheads.

\section{Evaluation} \label{Section: Evaluation}

In this section, we test full implementation on top of PyTorch~\cite{PyTorch} with an extensive set of evaluations.

\subsection{Experimental Setup} \label{Section: Evaluation Settings}


\textbf{Testbeds}. We choose four different mobile and embedded platforms imposing different architectural features to showcase the cross-platform nature of LegoDNN when it comes to hardware. We use Huawei Mate20 X, with 8 2.6GHz ARM-based cores and 8 GB memory; Xiaomi 5S Plus with 4 2.35GHz cores (Qualcomm Kryo CPU) and 4 GB memory; Raspberry Pi 4B with 4 1.5GHz Cortex-A72 cores (ARM v8) and 4 GB memory; and Jetson TX2 with 256-core NVIDIA Pascal GPU and 8 GB memory. The first two devices run Android 10.0 and the last two device runs Linux Ubuntu 18.04 LTS.

\textbf{DNN models and datasets, and applications}. To evaluate LegoDNN's generalization capability on different vision tasks, we selected two most important applications in mobile vision systems.
\begin{itemize}[noitemsep,nolistsep]
\item \textbf{Generic-category image recognition}. This type of vision tasks aims to recognize the generic category of an image. Without loss of generality, we select two popular DNN models (VGG16~\cite{VGG} and ResNet18~\cite{resnet}) and image recognition datasets (Cifar-10~\cite{Cifar10DataSet} and ImageNet2012~\cite{krizhevsky2012imagenet}). Cifar-10 and ImageNet2012 contain 60k and 1.28 million images and 10 and 1000 image categories, respectively, representing an easy and difficult vision tasks correspondingly.
    In image recognition, the inference accuracy is measured by \emph{top-1 classification accuracy} on the test set (10k and 50k test images in Cifar-10 and ImageNet2012 respectively).

\item \textbf{Real-time object detection and video analytic}. This type of vision tasks aims to detect objects in real-time with fast processing time while maintaining a required level of accuracy. We select the most commonly used model (YOLOv3~\cite{YOLO}) and a large-scale dataset (COCO2017~\cite{COCO}). COCO2017 contains 0.33 million images, 1.5 million objects, and 80 categories.
    In object detection, the inference accuracy is measured by \emph{mean average precision (mAP)}, which is a standard metric associated Microsoft COCO challenge. mAP@0.5 is used here to measure mAP over Intersection over Union (IoU) threshold 0.5.
\end{itemize}

\textbf{Multi-DNN scenarios}. We consider three different scenarios with different constraints on latency and accuracy: maximum accuracy, minimum latency, and balanced latency and accuracy. Each scenario includes a mix of different DNN models, which represent mobile vision systems that execute different image recognition and object detection tasks.
To test LegoDNN's adaptability to different loads, we consider three ranges of concurrently running applications: a small load of 1-6 applications, a medium load of 2-8 applications, and a large load of 3-10 applications.

\textbf{Compared baselines}. We implement and compare four state-of-the art model-grained scaling techniques from the literature:
(i) \emph{Filter pruning} identifies and removes filters and feature maps with small impacts on model accuracy~\cite{li2016pruning}. (ii) \emph{NestDNN} improves this technique by judging a filter as important if its extracted feature maps can differentiate images of different classes~\cite{NestDNN18}.
(iii) \emph{Low rank decomposition}~\cite{kim2015compression} transforms a DNN to a compressed model by ranking kernel tensors using Variational Bayesian Matrix Fatorization (VBMF) and decomposing tensors using Tucker.
(iv) Finally, \emph{knowledge distillation}~\cite{tian2019contrastive} employs the teacher model to generate feature maps that preserve information of training samples, and use these maps to training multiple descendant models.

We also compare LegoDNN with another category of techniques that dynamically scale FLOPs at run-time to trade off inference accuracy and latency:
(i) \emph{MSDNet}~\cite{msdnet} and Hardware-Aware Progressive Inference (HAPI)~\cite{laskaridis2020hapi} are designed for progressive inference. They contain several points that allow the early exist of inference when simple samples are correctly classified.
(ii) \emph{BlockDrop}~\cite{blockdrop} and \emph{ConvNet-AIG}~\cite{convnetaig} represent block-skipping solutions that dynamically skip irrelevant convolutional layers or blocks to reduce FLOPs.
(iii) \emph{CGNet}~\cite{cgnet} is a channel gating approach that dynamically identifies the parts of a DNN with negligible impacts on classification accuracy and skips these parts in inference.

\subsection{Evaluation of LegoDNN Components} \label{Section: Evaluation Settings}

\subsubsection{Evaluation of Offline Model/Block Generation}
\

This section's evaluation compares the model/block generation time of LegoDNN and four model-grained techniques.
In the baseline approaches, the re-training of descendant models takes most of the generation time.
In our approach, the generation time includes the generation, re-training, profiling time of descendant blocks.
For each DNN model, we generate five descendant models in the baseline approaches and five descendant blocks for each block in LegoDNN.
To make our comparisons fair, the re-training of each descendant model or block using the same training data, hyperparameter settings (the maximal epoch is 20), and platform (a GPU server with 48-GB Quadro RTX 8000 Graphics Card).
Table \ref{tab:trainingTime} lists the comparison results of five approaches.
We can see that LegoDNN uses the shortest generation time in all cases because its re-training of descendant blocks completes much faster than re-training descendant models in the baseline approaches.
In this evaluation, VGG16, YOLOv3, and ResNet18 have 5, 5, and 8 blocks, respectively. By just generating five descendant blocks for each block, LegoDNN provides 7,776 optional models for VGG16 and YOLOv3, and 1,679,616 models for ResNet18.
In contrast, the baseline approaches provide 6 optional model and they take at least several years to generate the same number of optional models with LegoDNN.

We note that with limited GPU memories, the applied scheduling policy also influences re-training time. We tested three policies: smallest block first, largest block first, and random order. The evaluation results show for VGG16, ResNet18, and YOLOv3, the training time in the second policy is 8.89, 2.07, and 6.14 times longer than that of the other policies. This is because training large blocks first reduces the number of concurrent training blocks and hence cannot adequately use computational resources in the GPU server.

\begin{table} [htp]
\caption{Benefit of LegoDNN on reducing model generation time (hour).}
\label{tab:trainingTime}
\begin{tabular}{|p{2.3cm}| p{1.3cm} |p{1.7cm}| p{1.3cm}|}
\hline
\multirow{1}{2.3cm}{Technique}   & \multirow{1}{1.2cm}{VGG16}     & ResNet18     & YOLOv3 \\ \hline

\multirow{1}{2.3cm}{Filter pruning} & 0.53&35&14 \\ \hline
\multirow{1}{2.3cm}{NestDNN} &0.95 &53 &34  \\ \hline
\multirow{2}{2.3cm}{Low rank decomposition} & \multirow{2}{2.3cm}{0.53} & \multirow{2}{2.3cm}{41}& \multirow{2}{2.3cm}{13} \\
 &  & &   \\ \hline
\multirow{1}{2.3cm}{Knowledge distillation} & \multirow{2}{2.3cm}{0.32} & \multirow{2}{2.3cm}{37} & \multirow{2}{2.3cm}{10.53} \\
 &  & &   \\ \hline
\multirow{1}{2.3cm}{\textbf{LegoDNN}} & \textbf{0.21} & \textbf{8.6}& \textbf{5} \\ \hline
\end{tabular}
\end{table}

\subsubsection{Evaluation of Accuracy/Latency Tradeoff}
\

This section's evaluation considers two runtime overheads that determine inference latency in mobile vision systems: available memory and computation complexity in terms of FLOPs~\cite{xu2019first}, and compares different techniques under the same latency constraints.

\begin{figure*}[htp]
\centering
  \includegraphics[scale=0.67]{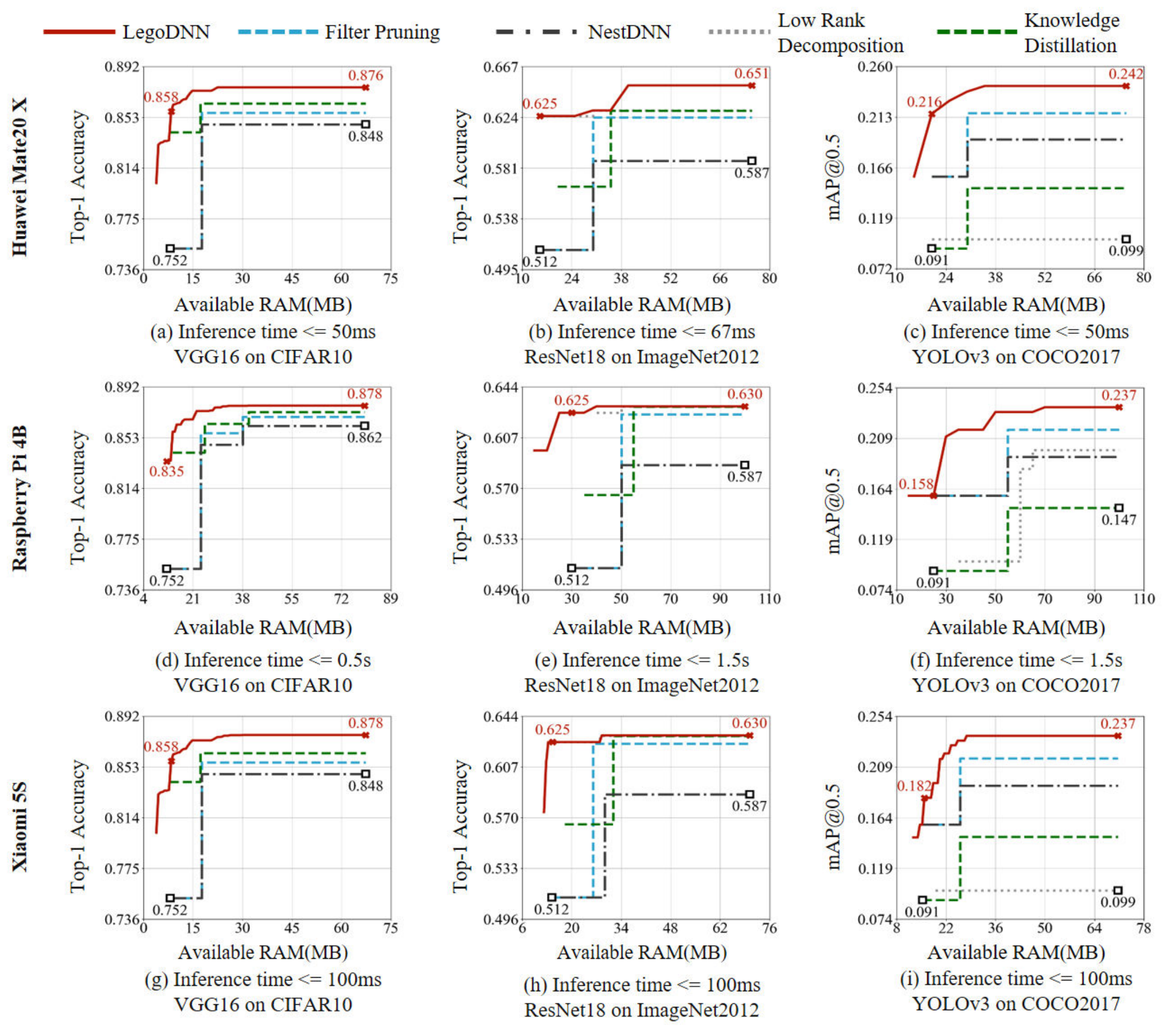}\\
  \caption{Comparison of inference accuracies between LegoDNN and model scaling approaches}
  \vspace{-2 mm} \label{Fig: ModelAccuarcy}
\end{figure*}

\textbf{Comparison to model-grained scaling}. In this evaluation, a DNN runs under different levels of memory availability, ranging from sufficient to stringent~\cite{mathur2017deepeye,lane2017squeezing}. The available memory is set according to an empirical study of 211 popular DL-based Android applications, which shows that a DNN usually occupies a few to dozens of MBs of memory resources in its execution~\cite{xu2019first}.
Figure \ref{Fig: ModelAccuarcy} illustrates the comparison between LegoDNN and baseline approaches across different DNNs and platforms. In LegoDNN, the stopping threshold is set to 0.005 and all scaling decisions are made within 1 second.
We have two key observations from the result.

First, LegoDNN consistently achieves higher accuracies than baseline approaches at every available resource. On average, LegoDNN increases accuracy by 31.74\%. The result indicates that our block-grained scaling is able to deliver state-of-the-art inference accuracy under a given memory budget. This is because achieves the largest memory utilizations in all cases (increasing utilization by 68.77\% in average), thus being able to use the largest model to increase inference accuracy.

Second, we observe that when the available memory becomes smaller, LegoDNN achieves more improvement in inference accuracy. On average, it achieves 52.66\% higher accuracy when the available memory is smaller than 30MB. This is because the block-grained scaler in LegoDNN is able to find the blocks having the largest influence on accuracy while first scaling down less important blocks. Small descendant models in LegoDNN hence benefit from these important blocks while other baseline models do not.

\textbf{Comparison to FLOP scaling techniques}. When considering time-sharing processors, scaling FLOPs provides another trade-off between inference latency and accuracy~\cite{msdnet,blockdrop,convnetaig,cgnet}.
Progressive inference techniques (MSDNet and HAPI) rely on specially designed DNNs. When comparing to them, our approach uses MobileNetv2~\cite{mobilenetv2} as the original DNN because it has similar FLOPs to the compared DNNs.
ResNet56~\cite{resnet} is used to test block-skipping (BlockDrop and ConvNet-AIG) and channel gating (CGNet) techniques, and our approach also uses it as the original DNN.

\begin{figure}[htp]
\centering
  \includegraphics[scale=0.5]{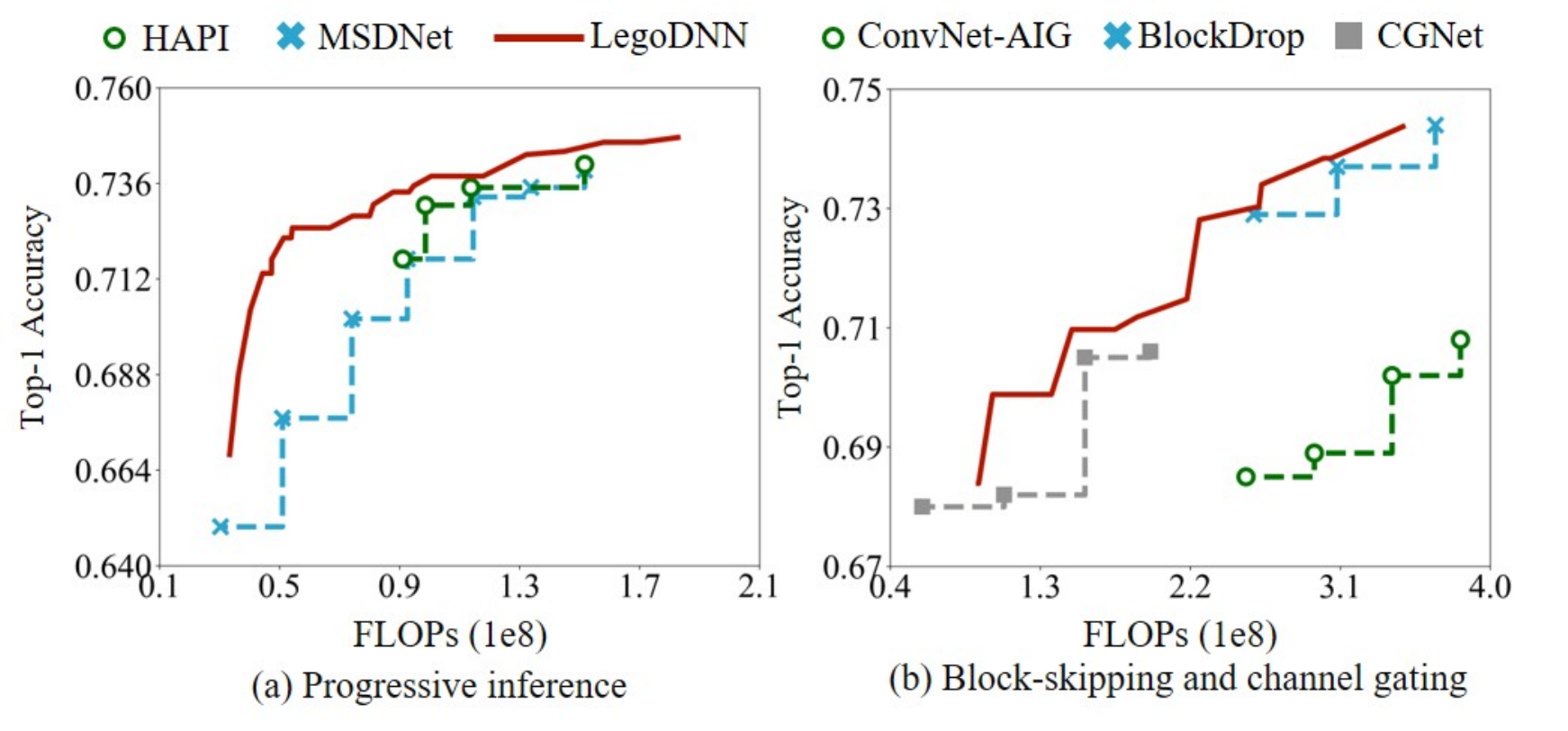}\\
  \caption{Comparison of inference accuracies between LegoDNN and FLOP scaling approaches}
  \label{Fig: FLOPs}
\end{figure}

Figure \ref{Fig: FLOPs} displays the comparison results on Cifar-100. We can see that our approach consistently achieves the highest accuracy when using the same FLOPs. This is because all FLOP scaling techniques skips parts of a DNN in inference and these parts may have large impact on accuracy. In contrast, LegoDNN reduces FLOPs by replacing large blocks with smaller ones, which can be viewed as an approximation of the large blocks and thus provide higher accuracies.

\begin{figure}[htp]
\centering
  \includegraphics[width=\columnwidth]{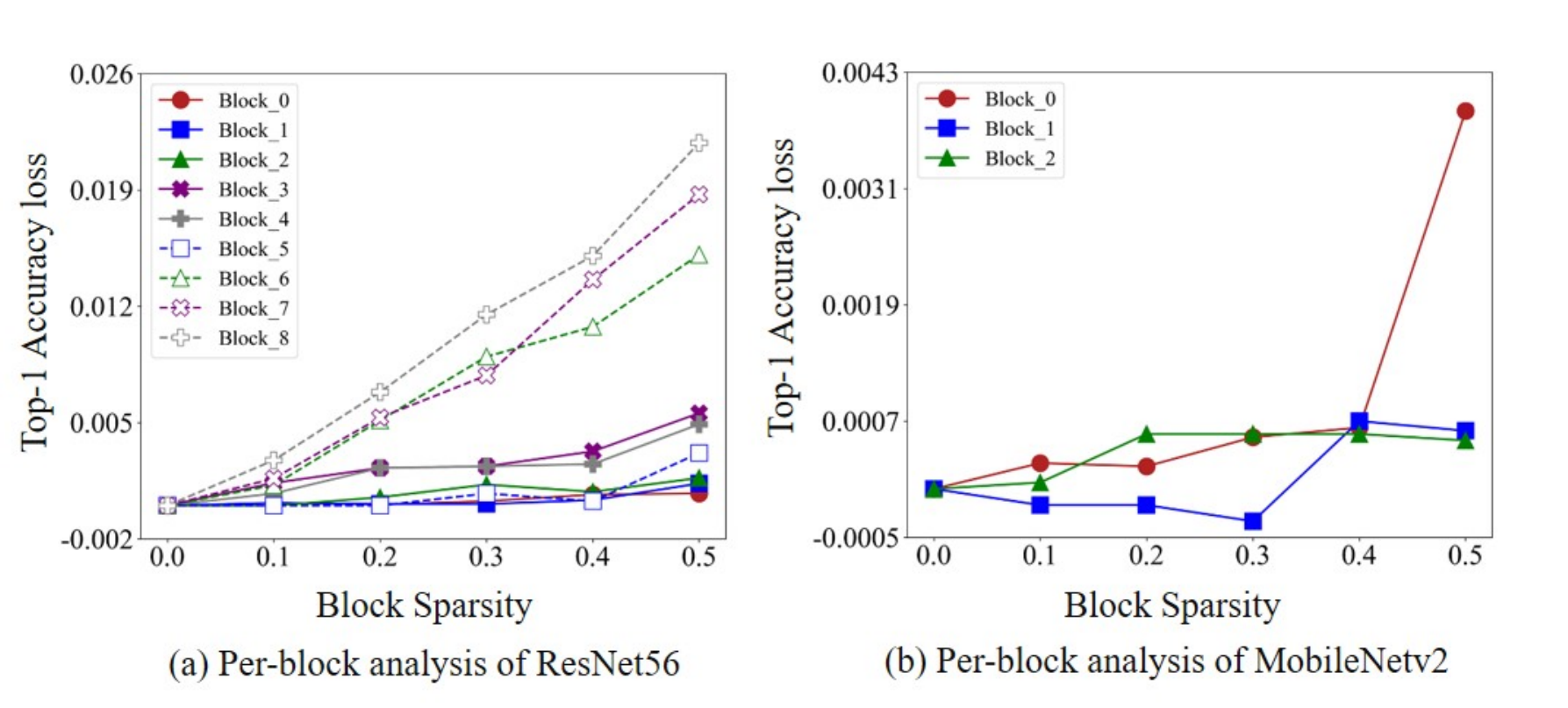}\\
  \caption{Per-block analysis of LegoDNN}
  \label{Fig: PerBlock}
\end{figure}

To support this claim, we demonstrate the per-block analysis of our approach when it is applied to ResNet56 and MobileNetv2 (the two original models used here). Figure \ref{Fig: PerBlock} shows the accuracy loss of each block under six model sparsities. We can see that even the blocks with the highest sparsity still have much smaller accuracy losses than those of FLOP scaling techniques (Figure \ref{Fig: FLOPs}).

\subsubsection{Evaluation of Scaling Optimization and Model Exchange}
\

This section's evaluation evaluate its two online scaling modules: scaling optimizer and model exchanger.

\textbf{Overheads and effectiveness of scaling optimization}. The optimization processing is controlled by the stopping threshold $\sigma$. This evaluation tests 6 values of $\sigma$ ranging from 5$e^{-6}$ to 1$e^{-2}$ on Raspberry Pi. For each value, 500 cases are tested and each case corresponds to the optimization of 10 DNNs of different latency and memory constraints.
The results show the optimization time ranges from 0.35 seconds to 0.20 seconds, and energy consumption ranges from 2.45 to 1.58 J on Raspberry Pi. This indicates a larger threshold needs smaller overheads, but it still achieves good optimization result: the difference to the optimal solution ranges from 3.97$e^{-9}$ to 3.7$e^{-4}$.

\begin{figure*}[htp]
\centering
  \includegraphics[scale=0.68]{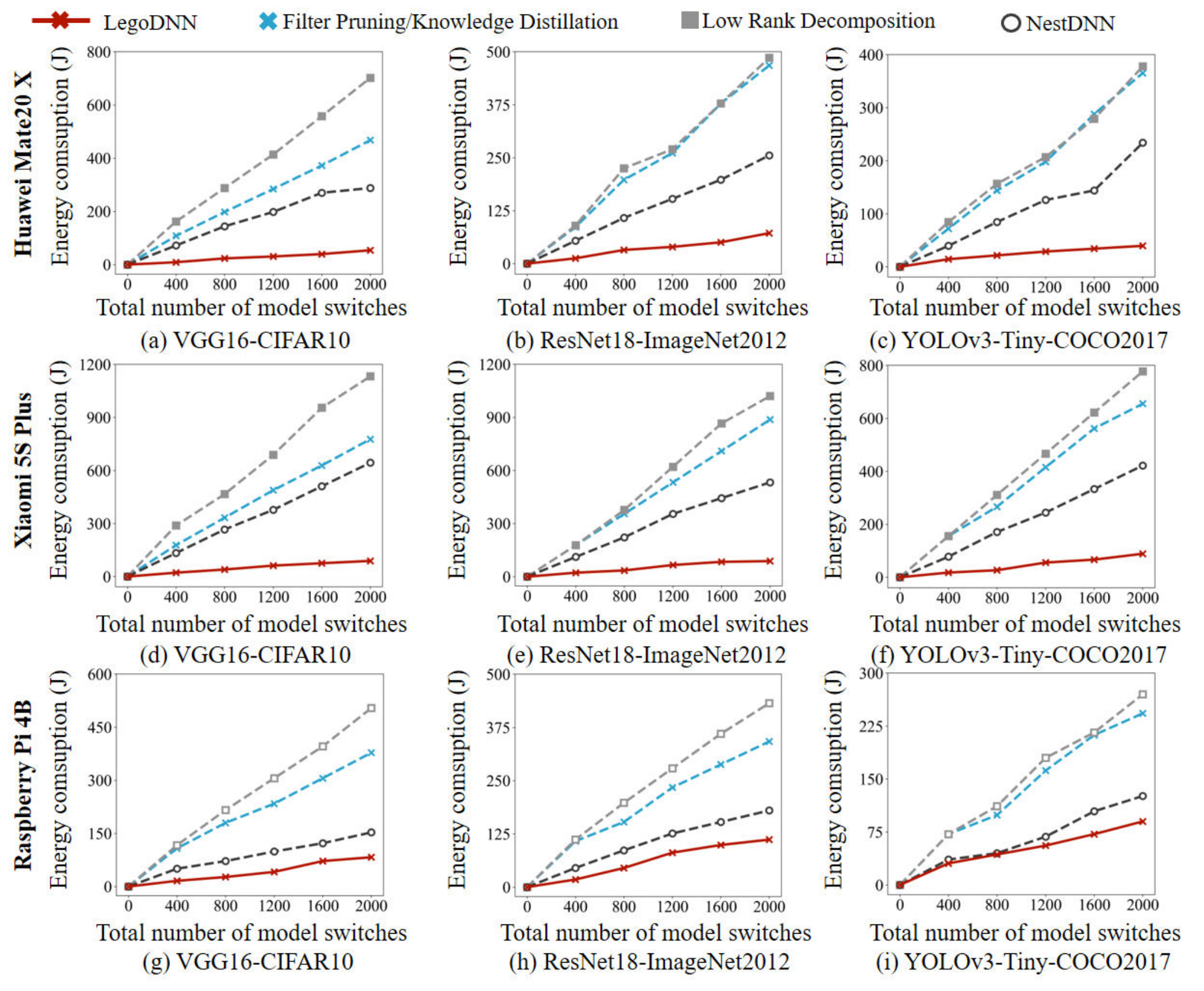}\\
  \caption{Comparison of energy consumptions in model scaling}
  \label{Fig: EnergyConsumptions}
\end{figure*}

\textbf{Overheads of model exchanger.} Another key metric of scaling is the energy consumption cased by model switching when the available resource changes. Figure \ref{Fig: EnergyConsumptions} shows the comparison results of the four approaches (the descendant models in filter puning and knowledge distillation have the same sizes) across three DNN models and three platforms. We used UNI-T UT658 power monitor to measure the energy consumptions.
As expected, energy consumed by switching the entire models in filter pruning and low rank decomposition is the largest. When considering all models and platforms, the page in and page out sizes are 3256.2 MB and 2799.8 MB in filter pruning and low rank decomposition respectively. In contrast, these sizes are 659.8 MB and 602 MB in NestDNN and LegoDNN respectively.
Compared to NestDNN, our approach still achieves much smaller energy consumptions because it needs no extra operations to change a model size.
The lightweight scheme in LegoDNN becomes more prominent when frequent scalings happen in resource and battery constrained mobile devices.
Overall, LegoDNN can save 100 to 1k J energy consumptions for every 2,000 model switches, and reduces scaling energies by 78.42\% on average on two smart phones, by 56.38\% on average on the embedded system, and by 71.07\% overall.

\subsection{Evaluation of Multi-DNN Scenario} \label{Section: Evaluation Settings}

\begin{figure}[htp]
\centering
  \includegraphics[scale=0.55]{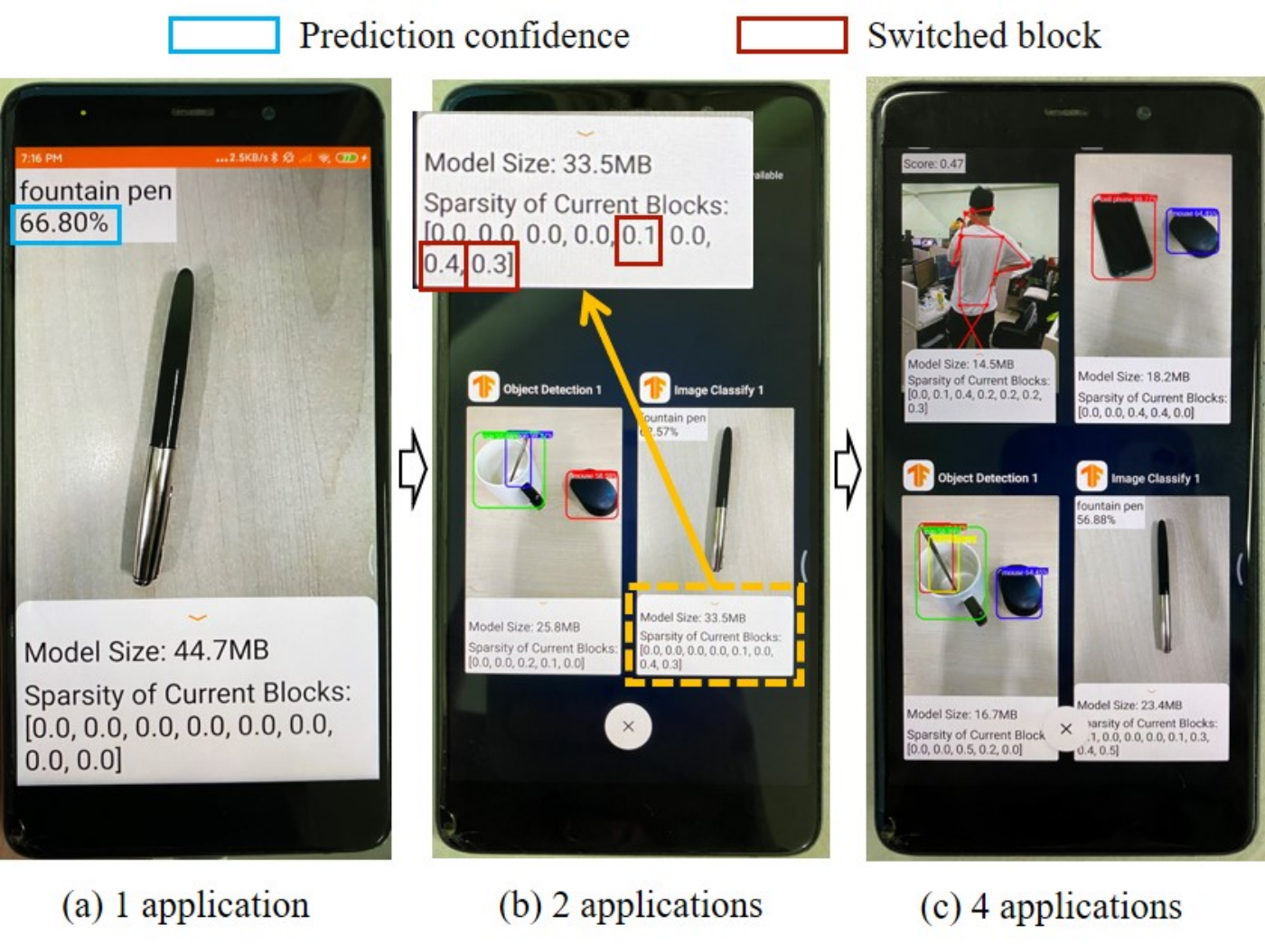}\\
  \caption{An example TensorFlow application}
  \label{Fig: TensorFlow}
\end{figure}

In this section, we evaluate multi-DNN scenarios on two real mobile devices: a CPU-based smart phone (Xiaomi 5S) and a GPU-based embedded device (Jetson TX2).
Figure \ref{Fig: TensorFlow} displays an example of TensorFlow Lite application~\cite{abadi2016tensorflow} on Xiaomi 5S.
In a multi-DNN scenario, an application's available resource decreases when more co-running applications exist (that is, from Figure \ref{Fig: TensorFlow}(a)'s 1 application to Figure \ref{Fig: TensorFlow}(c)'s 4 concurrent applications). LegoDNN provides a block-level scaling by just switching 3 blocks, thus still maintaining accurate models (high prediction confidences) with tighter resources.

In the following experiments, we developed a benchmark that runs a mix of DNN-driven vision tasks on Xiaomi 5S (using Android Debug Bridge (ADB)) or Jeston TX2.
At each 20 seconds of a 1,800-second test period, the benchmark creates a new application or kills a running application with certain probabilities (0.6 in this evaluation). This application is randomly selected from VGG16, ResNet18, and YOLOv3 following the settings of the previous section. This benchmark also allows the setting of the minimum and maximum numbers of concurrent application (e.g. 1 to 3 applications).

\subsubsection{Performance under Three Constraint Scenarios}
\

We first compare LegoDNN and the baseline approaches in the maximum accuracy and minimum latency scenarios. In the maximum accuracy scenario, all approaches have the same latency constraints: 2 seconds for Xiaomi 5S and 100 milliseconds for Jetson TX2.
In the minimal latency scenario, all approaches have the same accuracy constraints: the top-1 accuracies are 0.87 for VGG17, 0.685 for ResNet18, and the mAP is 0.27 for YOLOv3. The number of concurrent applications ranges from 3 to 5 on Xiaomi 5S and ranges from 7 to 10 on Jetson TX2.
We repeat each test 10 times and report the average inference latency or accuracy. Note that image recognition and object detection applications have different accuracy metrics, we thus use \emph{the percentage of loss in the average accuracy} of mixed applications, when comparing the evaluated approach against the accuracy of the original DNN model.

\begin{figure}[htp]
\centering
  \includegraphics[scale=0.46]{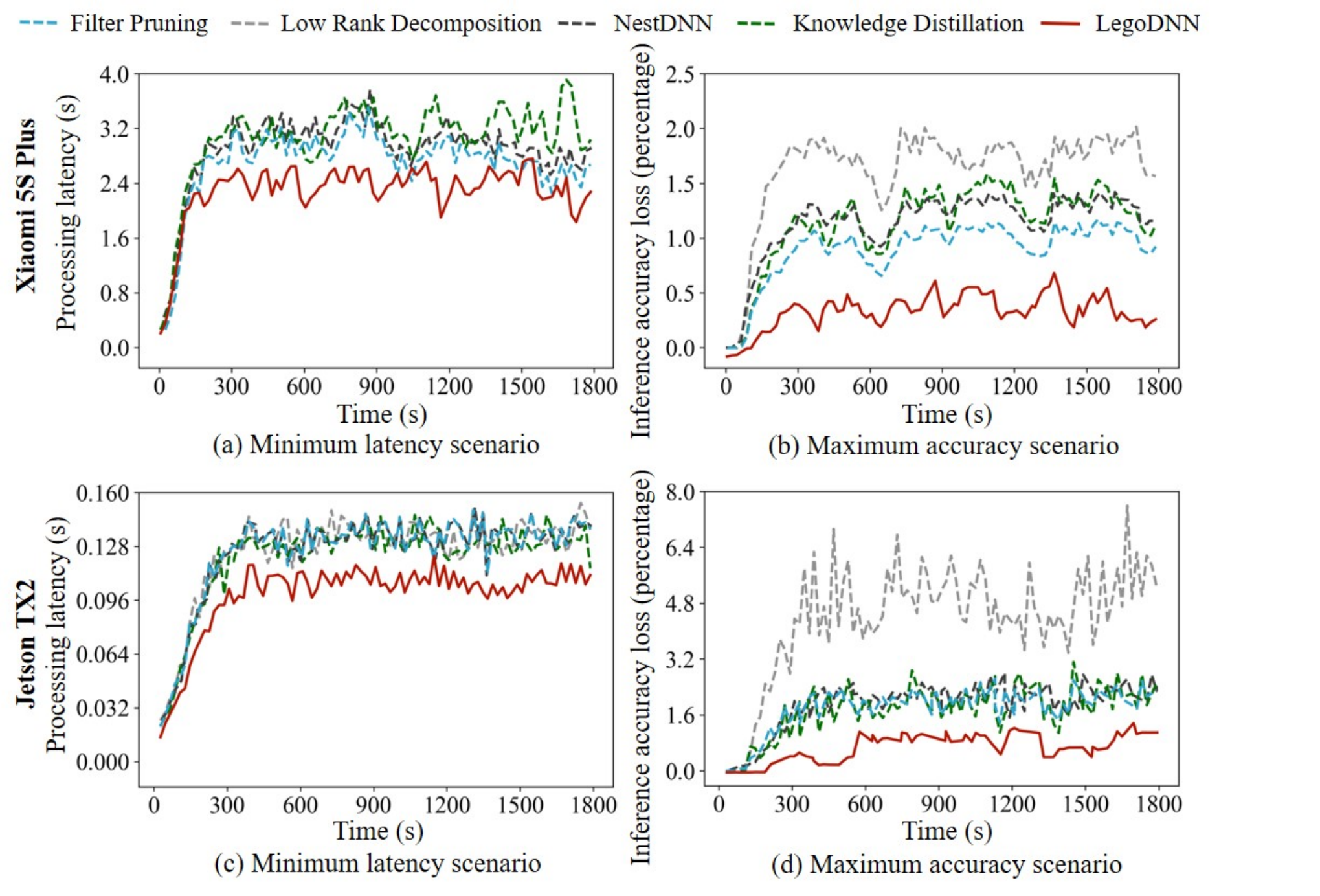}\\
  \caption{Performance comparison under maximum accuracy and minimum latency scenarios}
  \label{Fig: ConstrainedMinMax}
\end{figure}

\begin{figure}[htp]
\centering
  \includegraphics[scale=0.47]{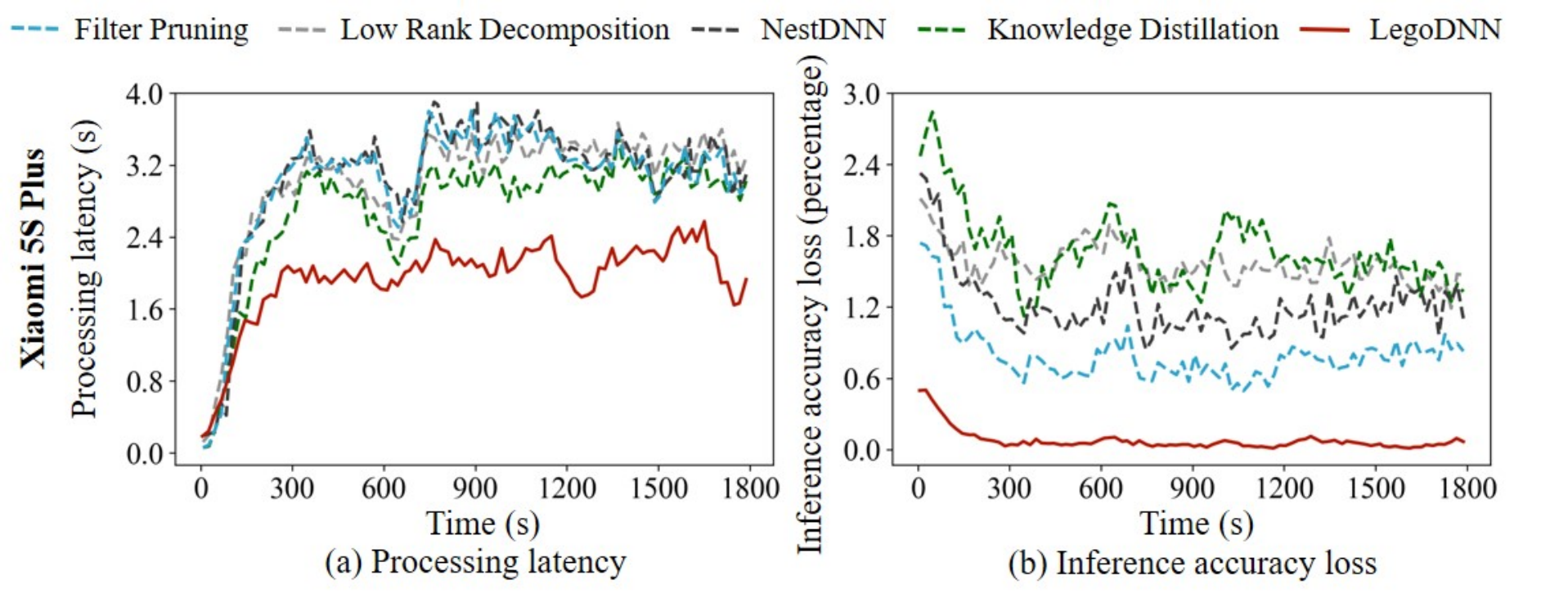}\\
  \caption{Performance comparison under the balanced latency and accuracy scenario}
  \label{Fig: BalancedSituation}
\end{figure}

Figure \ref{Fig: ConstrainedMinMax} shows the comparison result.
We can see when more and more concurrent applications run in the system (after test starts for 50 seconds), LegoDNN achieves considerable performance improvements against baseline approaches by delivering either lower latencies (minimal latency scenario) or smaller accuracy losses (maximum accuracy scenario). Overall, our approach achieves 21.62\% reductions (at most 2x) in latency and 69.71\% reductions in accuracy losses.
Figure \ref{Fig: BalancedSituation} further shows the comparison results of the balanced latency and accuracy scenario on Xiaomi 5S. Without any system constraints given, LegoDNN simultaneously achieves lower latencies and smaller accuracy losses compared to the baseline approaches. We have achieved consistent results on Jetson TX2.

\subsubsection{Performance under Different Loads}
\

\begin{figure}[htp]
\centering
  \includegraphics[scale=0.45]{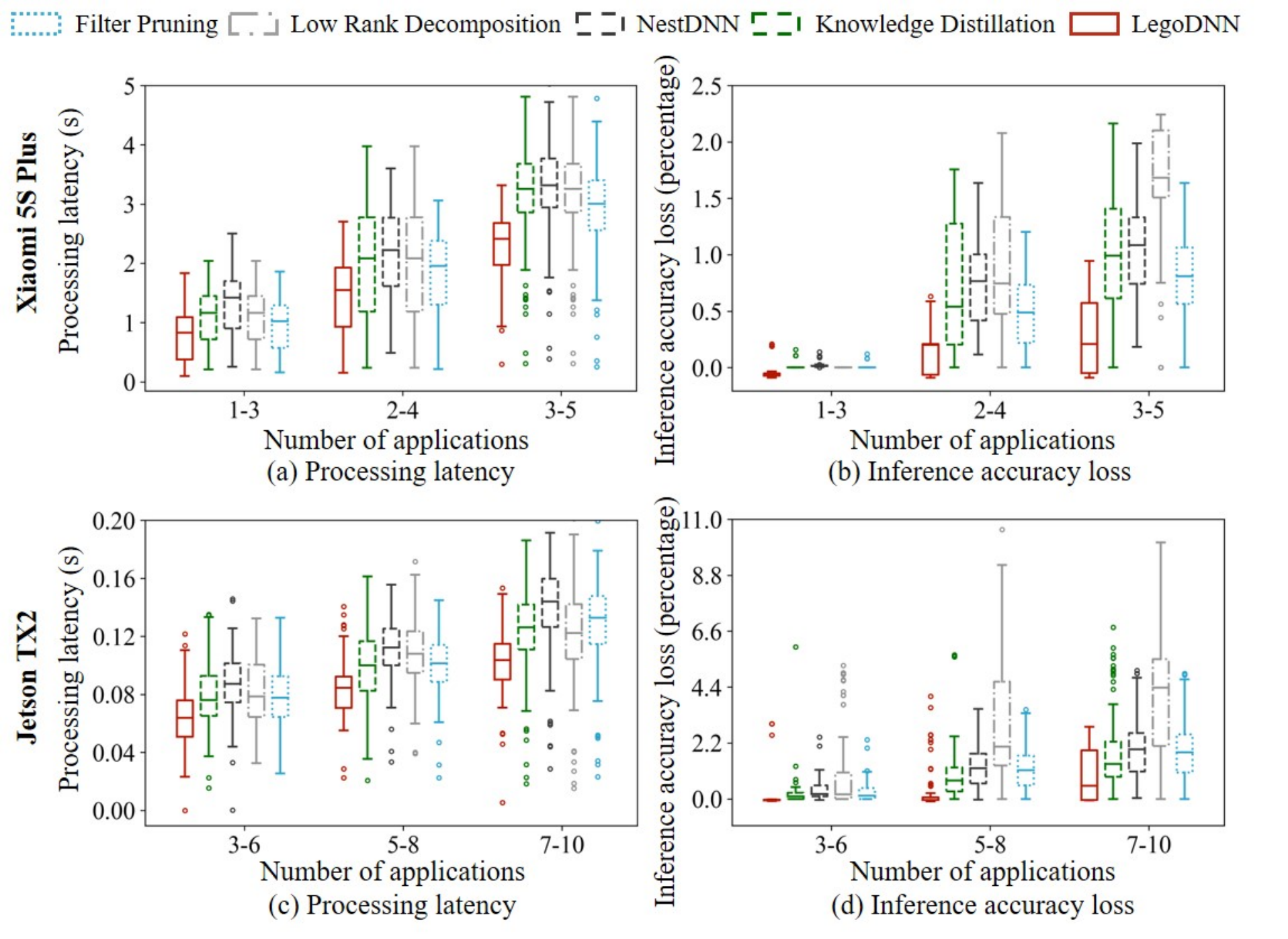}\\
  \caption{Comparison of latency and accuracy under different loads}
  \label{Fig: differentLoads}
\end{figure}

We extended the above evaluation by testing three different loads that represent small, medium, and large numbers of concurrent applications in mobile vision systems. Figure \ref{Fig: differentLoads} displays the comparison results in terms of the distributions of latencies and accuracy losses in the five approaches (using box plots). We can see that under different loads, LegoDNN still archives the lowest latencies and accuracy losses in all cases. On each platform, LegoDNN gains more performance improvements under larger loads. This means our approach performs better when systems are under heavy loads and their resources are saturated.

\section{Discussions}

\begin{figure*}[htp]
\centering
  \includegraphics[scale=0.63]{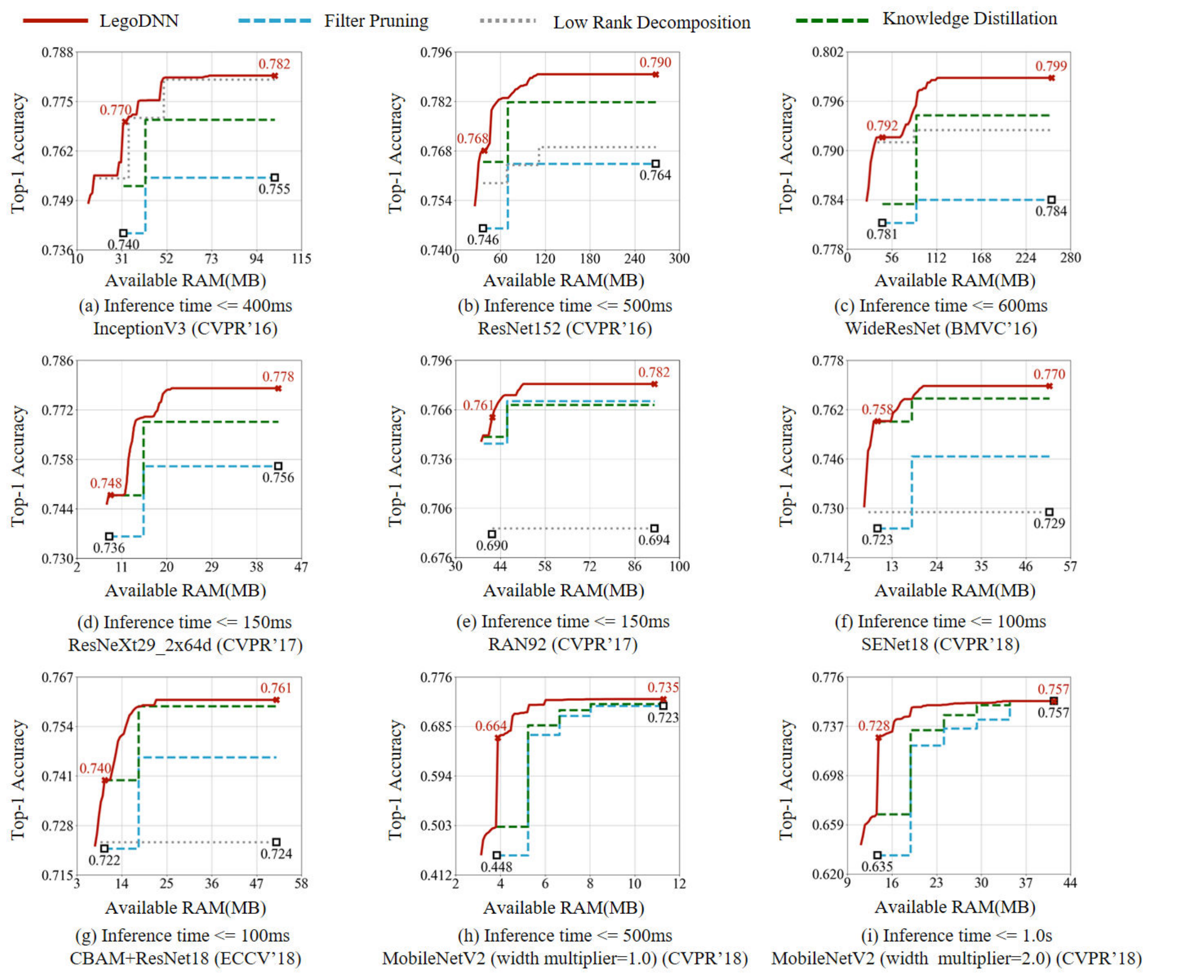}\\
  \caption{Applicability of LegoDNN to seven categories of DNNs}
   \vspace{-2 mm} \label{Fig: 9DNNs}
\end{figure*}

\textbf{Applicability of LegoDNN to DNNs}. LegoDNN represents the first framework that supports block-grained scaling of DNNs for mobile vision systems. LegoDNN can be generalized to support most of state-of-the-art DNNs~\cite{khan2020survey} of seven categories. We implemented and tested typical DNNs in each category: (1) spatial exploitation (VGG16~\cite{VGG}); (2) depth (ResNet152~\cite{resnet}); (3) multi-path (ResNet152~\cite{resnet}); (4) width (InceptionV3~\cite{inceptionv3}, ResNeXt29(2x64d)~\cite{resnext}, and WideResNet~\cite{wideresnet}); (5) feature map exploitation (SENet18~\cite{senet}); (6) attention (CBAM and ResNet18~\cite{cbam}, Residual Attention Network (RAN)~\cite{wang2017residual}); and (7) lightweight DNN (MobileNetV2 whose width multiplier is 1.0 and MobileNetV2 whose width multiplier is 2.0~\cite{mobilenetv2}). Figure \ref{Fig: 9DNNs} illustrates the comparison results between LegoDNN and the four baseline scaling techniques (NestDNN is inapplicable to complicated DNNs such as InceptionV3 and RAN) under the same latency constraints and using the Cifar-100 dataset. The results show that our approach consistently delivers higher inference accuracies for all 9 DNNs. Among all model-grained scaling techniques, knowledge distillation achieves the highest accuracies for most of the DNNs (the only exception is Inceptionv3). This means our approach can be integrated with knowledge distillation to further increase accuracies of descendant blocks.
We also note that LegoDNN is inapplicable when any filter in a DNN cannot be pruned. In the extreme case, each group convolution of the DNN only has one channel input, one map, and one filter that cannot be removed.

\begin{figure}[htp]
\centering
  \includegraphics[scale=0.49]{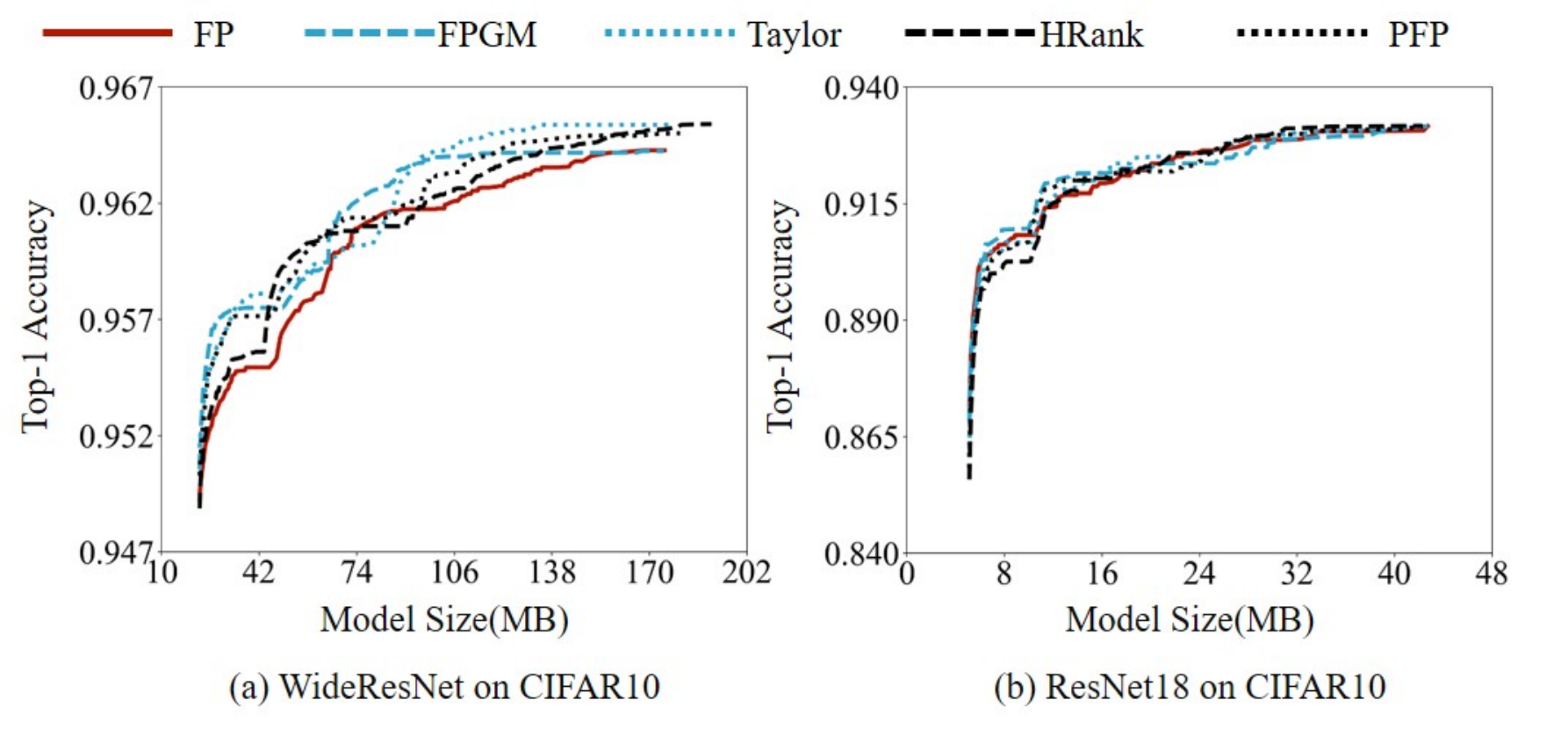}\\
  \caption{Implementation of LegoDNN using 5 model compression techniques}
   \label{Fig: modelCompression}
\end{figure}

\textbf{Discussion of other model compression techniques}. We implemented our approach using two types of model compression techniques: (1) \emph{data-independent} techniques: filter pruning~\cite{li2016pruning} and Filter Pruning via Geometric Median (FPGM)~\cite{he2019filter} that directly calculate the filters' importance and remove unimportant ones; (2) \emph{data-dependent} techniques: TaylorFOWeight~\cite{molchanov2019importance}, High-Rank Feature Map (HRank)~\cite{lin2020hrank}, and Provable Filter Pruning (PFP)~\cite{liebenwein2019provable}) that calculate the filters' importance based on training samples.
By testing ResNet18 and WideResNet as examples, Figure \ref{Fig: modelCompression} shows that all techniques achieves similar accuracies under different available memories. This is because although these techniques generate descendant blocks of different architectures, the blocks learn from the same original block in LegoDNN and hence have similar accuracies.

\begin{figure}[htp]
\centering
  \includegraphics[scale=0.5]{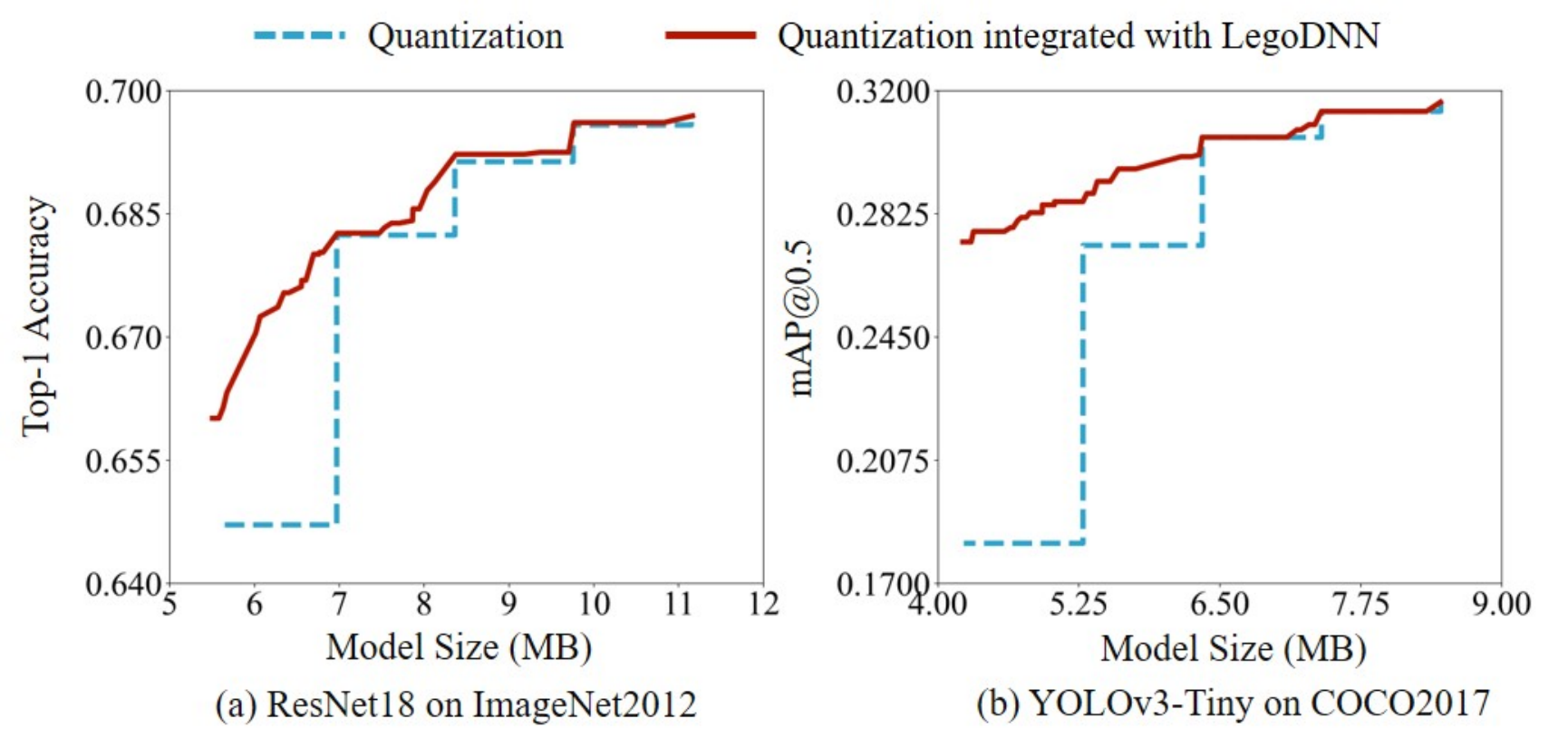}\\
  \caption{Integration of LegoDNN with a quantization technique}
  \label{Fig: quantization}
\end{figure}

In addition, we integrated our approach with a quantization technique~\cite{stock2019and}, which can considerably reduce model sizes but not necessarily reduce processing latency. In production mobile vision systems, quantization is usually used together with filter pruning techniques. The evaluation results in Figure \ref{Fig: quantization}(a) show the basic quantization technique transforms a DNN into several compressed models. When combining this technique with LegoDNN (Figure \ref{Fig: quantization}(b)), a large number of optional models are provided to enable fine-grained trade-offs between memory usage and accuracy.

\section{Related Work} \label{Section: Relatedwork}

There has been several recent works seeking to improve the performance of mobile vision applications at different stages of an inference job, including filtering of input data~\cite{chen2015glimpse,kang2017noscope,zhang2018ffs,canel2018picking,jain2018rexcam}, early exit of an inference task using intermediate output results~\cite{BranchyNet16,DistributedDNN2017,MobileEdge2018}, edge caching of inference results~\cite{chen2015glimpse,Cachier2017,Precog2017}, and model segmentation~\cite{li2018jalad,ko2018edge,kang2017neurosurgeon,MoDNN2017,MeDNN2017,DeepThings2018}.

\textbf{Model-grained DNN scaling}. Applying structured pruning techniques to trade off latency and model accuracy has been a hot topic in mobile vision systems~\cite{han2015deep,oh2018portable,li2018jalad,yang2017designing,reagen2016minerva} and related communities (e.g. causal speech enhancement~\cite{zhang2020multi}).
Existing scaling techniques~\cite{li2016pruning,NestDNN18,kim2015compression}) generate a limited number of descendant models due to the long model re-training time. At run-time, they dynamically select one of them according to the available resources.
However, the large gaps among these models may incur large performance degradations or accuracy losses.

\textbf{Scheduling of multiple DNNs}. Modern mobile vision systems also need to schedule the execution of multiple DNNs to optimize their latency, accuracy or energy~\cite{NestDNN18,bateni2020neuos,jiang2018mainstream,narayanan2018accelerating,han2016mcdnn,mathur2017deepeye}.
The proposed approach is not intended to replace, but rather complement the existing scheduling techniques: LegoDNN supports a block-grained trade-off between latency and accuracy and thus provides more options of model sizes in scheduling.

\textbf{Relationship to device-specific DNN generation and search}. Some recent techniques study the automatic generation of optimal DNNs for the underlying mobile devices~\cite{ZophL16,MnasNet2019}, or searching the optimal DNN for specific devices~\cite{kang2017noscope,Automating2019} or images~\cite{taylor2018adaptive}.
These techniques pre-generate DNN models according to the features of specific devices. In contrast, LegoDNN needs no prior knowledge when generating descendant blocks, and dynamically combines these blocks to adapt to the available device resources at run-time.

\section{Conclusion}

This paper presents the design, implementation and evaluation of LegoDNN, a framework that enables block-grained scaling of DNNs for mobile vision systems. Our approach can provide large exploration space of model size for a DNN by compressing its blocks offline, while optimally combining these blocks to provide high accuracy and low overheads at run-time. Extensive evaluation on complex scenarios against model compression and FLOP scaling algorithms prove the efficacy and practicality of LegoDNN. Our future work will explore the re-training of descendant blocks in the federated learning framework. Such re-training allows a descendant block to learn from multiple participants' DNNs to increase its accuracy and generalization.

\section*{Acknowledgments}
This work has been supported by the National Key Research and Development Plan of China (No. 2018YFB1003701), and the National Natural Science Foundation of China (No. 61872337). Corresponding author: Chi Harold Liu.

\bibliographystyle{plain}
\bibliography{references}

\end{document}